\documentclass{article} 
\usepackage{iclr2017_conference,times}
\usepackage{hyperref}
\usepackage{url}
\usepackage{graphicx}
\usepackage{booktabs}       
\usepackage{amsfonts}       
\usepackage{amsmath}       
\usepackage{algorithm}
\usepackage{algorithmic}
\usepackage{color}

\title{Deep unsupervised learning through spatial contrasting}

\author{
  Elad Hoffer\\
  Technion - Israel Institute of Technology\\
  Haifa, Israel\\
  \texttt{ehoffer@tx.technion.ac.il} \\
    \And
  Itay Hubara\\
  Technion - Israel Institute of Technology\\
  Haifa, Israel\\
  \texttt{itayh@tx.technion.ac.il} \\
      \And
  Nir Ailon \thanks{The author acknowledges the generous support of ISF grant number 1271/13}\\
  Technion - Israel Institute of Technology\\
  Haifa, Israel\\
  \texttt{nailon@cs.technion.ac.il}\\
}

\begin{document}

\maketitle

\begin{abstract}
Convolutional networks have marked their place over the last few years as the best performing model for various visual tasks. 
They are, however, most suited for supervised learning from large amounts of labeled data. 
Previous attempts have been made to use unlabeled data to improve model performance by applying unsupervised techniques. These attempts require different architectures and training methods. In this work we present a novel approach for unsupervised training of
Convolutional networks that is based on contrasting between spatial regions within images. 
This criterion can be employed within conventional neural networks and trained using standard techniques such as  SGD and back-propagation, thus complementing supervised methods.
\end{abstract}

\section{Introduction}
For the past few years convolutional networks (ConvNets, CNNs)  \citet{lecun1998gradient}  have proven themselves as a successful model
for vision related tasks \citet{Krizhevsky2012} \citet{mnih2015human} \citet{pinheiro2015learning} \citet{razavian2014cnn}. 
A convolutional network is composed of multiple convolutional and pooling layers,
followed by a fully-connected affine transformations. As with other neural network models, each layer is typically followed by
a non-linearity transformation such as a rectified-linear unit (ReLU). \\
A convolutional layer is applied by cross correlating an image with a trainable weight filter. This stems from the
assumption of stationarity in natural images, which means that  features learned for one local region in an image can be shared for other regions and images. 

Deep learning models, including convolutional networks, are usually trained in a supervised manner, requiring large amounts
of labeled data (ranging between thousands to millions of examples per-class for classification tasks) in almost all modern applications. 
These models are optimized a variant of stochastic-gradient-descent (SGD) over batches of images sampled from the
whole training dataset and their ground truth-labels. Gradient estimation for each one of the optimized parameters is done by back propagating the
objective error from the final layer towards the input. This is commonly known as "backpropagation" \citet{rumelhart1988learning}.

One early well known usage of unsupervised training of deep architectures was as part of a pre-training procedure used for obtaining an effective initial state
of the model. The network was later fine-tuned in a supervised manner as displayed by \citet{hinton2007recognize}.  
Such unsupervised pre-training procedures were later abandoned, since they provided no apparent benefit over other initialization heuristics in 
more careful fully supervised training regimes. This led to the de-facto almost exclusive usage of neural networks in supervised environments.

In this work we will present a novel unsupervised learning criterion for convolutional network based on comparison of features extracted from regions within images.
Our experiments indicate that by using this criterion to pre-train networks we can improve their performance and achieve state-of-the-art results.
\section{Previous works}
Using unsupervised methods to improve performance have been the holy grail of deep learning for the last couple of years and vast research efforts have been focused on that.
We hereby give a short overview of the most popular and recent methods that tried to tackle this problem.
\paragraph{AutoEncoders and reconstruction loss}
These are probably the most popular models for unsupervised learning using neural networks, and ConvNets in particular. Autoencoders are NNs which aim to transform inputs into outputs with the least possible amount of distortion. An Autoencoder is constructed
using an encoder $G(x;w_1)$  that maps an input to a hidden compressed representation, followed by a decoder $F(y;w_2)$, 
that maps the representation back into the input space. Mathematically, this can be written in the following general form:
\begin{align*}
&\hat{x} = F(G(x;w_1);w_2)
\end{align*}
The underlying encoder and decoder contain a set of trainable parameters that can be tied together and optimized for a predefined criterion. The encoder and decoder can have different architectures, including fully-connected neural networks, ConvNets and others. The criterion used for training is the reconstruction loss, usually the mean squared error (MSE) between the original input and its reconstruction \cite{zeiler2010deconvolutional}
\begin{align*}
&min \|x-\hat{x}\|^2 \\
\end{align*}
This allows an efficient training procedure using the aforementioned backpropagation and SGD techniques. Over the years autoencoders gained fundamental role in unsupervised learning and many modification to the classic architecture were made. \cite{ng2011sparse} regularized the latent representation to be sparse, \cite{vincent2008extracting} substituted the input with a noisy version thereof, requiring the model to denoise while reconstructing. Kingma et al. (2014) obtained very promising results with variational autoencoders (VAE). A variational autoencoder model inherits typical autoencoder architecture, but makes strong assumptions concerning the distribution of latent variables. They use variational approach for latent representation learning, which results in an additional loss component and specific training algorithm called Stochastic Gradient Variational Bayes (SGVB). VAE assumes that the data is generated by a directed graphical model $p(x|z)$ and require  the encoder to learn an approximation $q_{w_1}(z|x)$ to the posterior distribution $p_{w_2}(z|x)$ where $w_1$ and $w_2$ denote the parameters of the encoder and decoder. 
The objective of the variational autoencoder in this case has the following form:
\begin{align*}
&\mathcal{L}(w_1,w_2,x)=-D_{KL}(q_{w_1}(z|x)||p_{w_2}(z))+\mathbb{E}_{q_{w_1}(z|x)}\big(\log p_{w_2}(x|z)\big)
 \\
 \end{align*}
Recently, a stacked set of denoising autoencoders architectures showed promising results in both semi-supervised and unsupervised tasks. A stacked what-where autoencoder by \cite{whatwhereae} computes a set of complementary variables that enable reconstruction whenever a layer implements a many-to-one mapping. Ladder networks by \cite{rasmus2015semi} - use lateral connections to allow higher levels of an autoencoder to focus on
invariant abstract features by applying a layer-wise cost function.

\paragraph{Exemplar Networks:}
The unsupervised method introduced by\cite{dosovitskiy2014discriminative} takes a different approach to this task and trains the network to discriminate between a set of pseudo-classes.
Each pseudo-class is formed by applying multiple transformations to a randomly sampled image patch. 
The number of pseudo-classes can be as big as the size of the input samples. 
This criterion ensures that different input samples would be distinguished while providing robustness to the applied transformations.

\paragraph{Context prediction}
Another method for unsupervised learning by context was introduced by \citet{doersch2015unsupervised}. This method uses an auxiliary
criterion of predicting the location of an image patch given another from the same image. This is done by classification to 1 of 9 possible locations.

\paragraph{Adversarial Generative Models:}
This a recently introduced model that can be used in an unsupervised fashion \cite{goodfellow2014generative}. Adversarial Generative Models uses a set of
networks, one trained to discriminate between data sampled from the true underlying distribution (e.g., a set of images), and a separate
generative network trained to be an adversary trying to confuse the first network.   
By propagating the gradient through the paired networks,
the model learns to generate samples that are distributed similarly to the source data. As shown by \cite{radford2015unsupervised},this model can create useful latent representations for subsequent classification tasks as demonstrated 

\paragraph{Sampling Methods:}
Methods for training models to discriminate between a very large number of classes often use a \emph{noise contrasting criterion}.
In these methods, roughly speaking, the posterior probability $P(t|y_t)$ of the ground-truth target $t$ given the model output
on an input sampled from the true distribution $y_t=F(x)$ is maximized,
while the probability $P(t|y_n)$ given a noise measurement $y=F(n)$ is minimized. This was successfully used in a language domain
to learn unsupervised representation of words.  The most noteworthy case is the word2vec model introduced by \cite{word2vec}.
When using this setting in language applications, a natural contrasting noise is a smooth approximation of the Unigram distribution. A suitable
contrasting distribution is less obvious when data points are sampled from a high dimensional continuous space, such as in the case of patches of images.

\subsection{Problems with Current Approaches}
Only recently the potential of ConvNets in an unsupervised environment began to bear fruit, still we believe it is not fully uncovered. 


The majority of unsupervised optimization criteria currently used are based on variations of reconstruction losses.  
One  limitation of this fact is that a pixel level reconstruction is non-compliant with the idea  of a discriminative objective, 
which is expected to be agnostic to low level information in the input.
In addition, it is evident that MSE is not best suited as a measurement to
compare images, for example, viewing the possibly large square-error between an image and a single pixel shifted copy of it. 
Another problem  with recent approaches such as \citet{rasmus2015semi,zeiler2010deconvolutional}  is their need to extensively modify the original convolutional network model.
This leads to a gap between unsupervised method and the state-of-the-art, supervised, models for classification - which can hurt future attempt to reconcile them in a unified framework, and also to efficiently leverage unlabeled data with otherwise supervised regimes.

\section{Learning by Comparisons}
The most common way to train NN is by defining a loss function between the target values and the network output. Learning by comparison approaches the supervised task from a different angle. The main idea is to use distance comparisons between samples to learn useful representations. For example, we consider relative and qualitative examples of the form “$X_1$ is closer to $X_2$ than $X_1$ is to $X_3$. Using a comparative measure with neural network to learn embedding space was introduced in the ``Siamese network'' framework by \citet{bromley1993signature} and later used in the works of \citet{chopra2005learning}.
One use for this methods is when the number of classes is  too large or expected to vary over time, as in the case
of face verification, where a face contained in an image has to compared against another image of a face. 
This problem was
recently tackled by \citet{schroff2015facenet} for training a convolutional network model on triplets of examples.   There, one image served as an \emph{anchor} $x$, and
an additional pair of  images served as a positive example $x_{+}$  (containing an instance of the  face of the same person) together with a negative example $x_-$, containing a face of a different person.
The training objective  was on the embedded distance of the input faces, where the distance between the anchor and positive example
is adjusted to be smaller by at least some constant $\alpha$ from the negative distance. More precisely, the loss function used in this case
was defined as  
\begin{equation}
\label{margin_distance}
L(x,x_{+},x_{-}) = \max{\left\{\|F(x)-F(x_{+})\|_2-\|F(x)-F(x_{-})\|_2 + \alpha, 0\right\}}
\end{equation}
where $F(x)$ is the embedding (the output of a convolutional neural network), and $\alpha$ is a predefined margin constant.
Another similar model used by \citet{hoffer2015deep} with triplets comparisons for classification, where examples from the same class were trained
to have a lower embedded distance than that of two images from distinct classes. This work introduced a concept of a distance ratio loss,
where the defined measure amounted to:
\begin{equation}
\label{ratio_distance}
 L(x,x_{+},x_{-}) =-\log \frac{e^{-\|F(x)-F(x_{+})\|_2}}{e^{-\|F(x)-F(x_{+})\|_2}+{e^{-\|F(x)-F(x_{-})\|_2}}}\ 
\end{equation}
This loss has a flavor of a probability of a biased coin flip.  By `pushing' this probability to zero, we express the
objective that pairs of samples coming from distinct classes should be less similar to each other, compared to pairs
of samples coming from the same class. 
It was shown empirical by \citet{balntas2016pn} to provide better feature embeddings than the margin based distance loss \ref{margin_distance}

\section{Our Contribution: Spatial Contrasting}
One implicit assumption in convolutional networks, is that features 
are gradually learned hierarchically, each level in the hierarchy corresponding to a layer in the network.
Each spatial location within a layer corresponds to a
region in the original image.  It is empirically observed that deeper layers tend to contain more `abstract' information from the image. 
Intuitively,
features describing 
different regions within the same image are likely to be  semantically similar (e.g.  different parts of an animal), and
indeed the corresponding deep representations tend to be similar.
Conversely, regions from two probably unrelated images (say, two images chosen at
random) tend to be far from each other in the deep representation.
This logic is commonly used in modern deep networks such as \citet{inception} \citet{nin} \citet{res}, 
where a global average pooling is used to aggregate spatial features in the final layer used for classification. 

Our suggestion is that this property, often observed as a side effect of supervised applications, can be used as a desired objective when learning deep representations
 in an unsupervised task.  Later, the resulting representation can be used, as typically done, as a starting point
or a supervised learning task.
We call this idea which we formalize below \emph{Spatial contrasting}.  
The spatial contrasting criterion is similar to noise contrasting estimation \cite{nce1} \cite{nce2}, in trying to train a model by
maximizing the expected probability on desired inputs, while minimizing it on contrasting sampled measurements.

\subsection{Formulation}
We will concern ourselves with samples of images patches $\tilde{x}^{(m)}$ taken from an image $x$. Our convolutional network model, denoted by $F(x)$, extracts spatial features $f$ so that $f^{(m)}=F(\tilde{x}^{(m)})$ for an image patch $\tilde{x}^{(m)}$. We wish to optimize our
model such that for two features representing patches taken from the same image $\tilde{x}_i^{(1)},\tilde{x}_i^{(2)}\in x_i$
for which $f_i^{(1)}=F(\tilde{x}_i^{(1)})$ and $f_i^{(2)}=F(\tilde{x}_i^{(2)})$, 
the conditional probability $P(f_i^{(1)}|f_i^{(2)})$ will be maximized. \\
This means that features from a patch taken from a specific image  can  effectively predict,
under our model, features extracted from other patches in the same image.  Conversely, we want our model to minimize $P(f_i|f_j)$ for $i,j$ being two patches taken
from distinct images.
Following the logic presented before, we will need to sample
\emph{contrasting patch} $\tilde{x}^{(1)}_j$ from a different image $x_j$ such that $P(f_i^{(1)}|f_i^{(2)})>P(f_j^{(1)}|f_i^{(2)})$, where $f_j^{(1)}=F(\tilde{x}^{(1)}_j)$. 
In order to obtain contrasting samples,
we use regions from two random images in the training set.
We will use a distance ratio, described earlier \ref{ratio_distance} for the supervised case, to represent the probability two feature vectors were taken from the same image.
The resulting training loss for a pair of  images will be defined as

\begin{equation}
 L_{SC}(x_1,x_2) = -\log \frac{e^{-\|f_1^{(1)} - f_1^{(2)}\|_2}}{e^{-\|f_1^{(1)} - f_1^{(2)}\|_2} + e^{-\|f_1^{(1)} - f_2^{(1)}\|_2}}
\end{equation}

\begin{figure}[h]
  \centering
  \label{sc_drawing}
\includegraphics[width=0.4\textwidth]{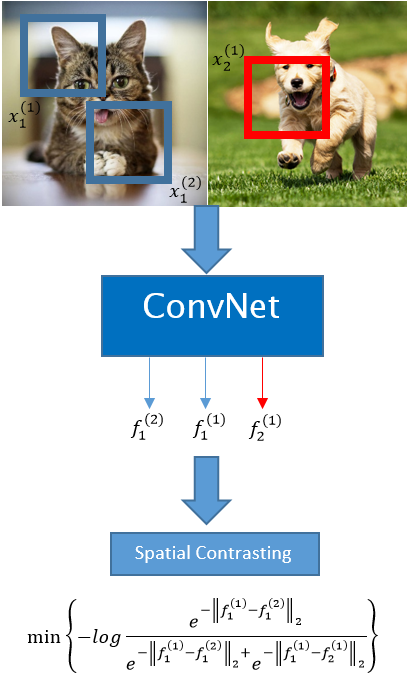}
  \caption{Spatial contrasting depiction.}
\end{figure}

Effectively minimizing a log-probability under the SoftMax measure.
This formulation is portrayed in figure \ref{sc_drawing}. 
Since we sample our contrasting sample from the same underlying distribution, we can evaluate this loss considering the image patch as both
patch compared (anchor) and contrast symmetrically. The final loss will be the average between these estimations:
$$\hat{L_{SC}}(x_1,x_2) = \frac{1}{2}\left[L_{SC}(x_1,x_2) + L_{SC}(x_2,x_1) \right]$$

\subsection{Method}
Since training convolutional network is done in batches of images, we can use the multiple samples in each batch to train our model. Each image serves as a source for
both an anchor and positive patches, for which the corresponding features should be closer, and also a source for contrasting samples for all the other images in that batch.
For a batch of $N$ images, two samples from each image are taken, and $N^2$ 
different distance comparisons are made. The final loss is the average distance ratio
for images in the batch:
\begin{equation}
\bar{L_{SC}}(\{x\}_{i=1}^N) = \frac{1}{N}\sum_{i=1}^N L_{SC}(x_i,\{x\}_{j\ne i}) = -\frac{1}{N}\sum_{i=1}^N \log \frac{e^{-\|f_i^{(1)} - f_i^{(2)}\|_2}}{\sum_{j=1}^N e^{-\|f_i^{(1)} - f_j^{(2)}\|_2}}
\end{equation}
Since the criterion is differentiable with respect to its inputs, it is fully compliant with standard methods for training convolutional network and specifically using
backpropagation and gradient descent.
Furthermore, SC can be applied to any layer in the network hierarchy.
In fact, SC can be used at  multiple layers within 
the same convolutional network.
The spatial properties of the features means that we can also sample from feature space $\tilde{f}^{(m)}\in f$ instead of from the original image, which we use to simplify implementation.
The complete algorithm for batch training is described in \ref{alg}. 
This algorithm is also related to the batch normalization layer \citet{ioffe2015batch}, a recent usage for batch statistics in neural networks. 
Spatial contrasting also uses the batch statistics, but to sample contrasting patches.

\begin{algorithm}
\caption{Calculation the spatial contrasting loss}
\label{alg}
\begin{algorithmic}
\REQUIRE{$X=\{x\}_{i=1}^N$} \# Training on batches of images\\
 \vspace{\baselineskip}
\# Get the spatial features for the whole batch of images \\
\# Size: $N \times W_f \times H_f \times C $

\STATE $\{f\}_{i=1}^N \leftarrow \mathtt{ConvNet}(X)$ \\

\vspace{\baselineskip}
\# Sample spatial features and calculate embedded distance between all pairs of images \\

\FOR{i = 1 \TO N}
	\STATE $\tilde{f}_i^{(1)} \leftarrow \mathtt{sample}(f_i)$
	\FOR{j = 1 \TO N}
	  \STATE $\tilde{f}_j^{(2)} \leftarrow \mathtt{sample}(f_j)$
	  \STATE $Dist(i,j) \leftarrow \|\tilde{f}_i^{(1)} - \tilde{f}_j^{(2)}\|_2$
	\ENDFOR
\ENDFOR

\vspace{\baselineskip}
\# Calculate log SoftMax normalized distances\\
$d_i \leftarrow -\log \frac{e^{-Dist(i,i)}}{\sum_{k=1}^N e^{-Dist(i,k)}}$

\vspace{\baselineskip}
\# Spatial contrasting loss is the mean of distance ratios\\
\RETURN $\frac{1}{N}\sum_{i=1}^N d_i$
\end{algorithmic}
\end{algorithm}

\section{Experiments}
In this section we report empirical results showing that using SC loss as an unsupervised pretraining procedure 
can improve state-of-the-art performance on subsequent classification. 
We experimented with MNIST,
CIFAR-10 and STL10 datasets. We used modified versions of well studied networks such as those of \citet{nin} \citet{rasmus2015semi}.
A detailed description of our architecture can be found in Appendix A.

In each one of the experiments, we used the spatial contrasting criterion to train the network on the unlabeled images. Training was done by using SGD with an initial learning rate
of $0.1$ that was decreased by a factor of $10$ whenever the measured loss stopped decreasing. After convergence,
 we used the trained model as an initialization
for a supervised training on the complete labeled dataset. The supervised training was done following the same regime, only starting with a lower initial learning rate of $0.01$. 
We used mild data augmentations, such as small translations and horizontal mirroring.\\
The datasets we used are:
\begin{itemize}
\item{STL10} (\citet{stl10}).  This dataset consists of $100,000 \ 96\times 96$ colored, unlabeled images, together with 
another set of $5,000$ labeled training images and $8,000$ test images .  The label space consists of 10 object classes. 
\item{Cifar10} (\citet{krizhevsky2009learning}). The well known CIFAR-10 is an image classification benchmark dataset containing $50,000$ training images and $10,000$ test images.  The image sizes  $32 \times 32$ pixels, with color.  The classes are
airplanes, automobiles, birds, cats, deer, dogs, frogs, horses, ships and trucks. 
\item{MNIST}  (\citet{lecun1998gradient}).  The MNIST database of handwritten digits is one of the most studied dataset benchmark
for image classification. The dataset contains 60,000 examples of handwritten digits from 0 to 9 for training and 10,000 additional  examples for testing. 
Each sample is a 28 x 28 pixel gray level image. 
\end{itemize}
\subsection{Results on STL10}
Since STL10 is comprised of mostly unlabeled data, it is the most suitable to highlight the benefits of the spatial contrasting criterion. The initial training was unsupervised, as described
earlier, using the entire set of $105,000$ samples (union of the original unlabeled set and labeled training set).
The representation outputted by the training, was used to initialize supervised training on the $5,000$ labeled images. Evaluation was done 
on a separate test set of $8,000$ samples. Comparing
with state of the art results \ref{stl10_table} we see an improvement of ~7\% in test accuracy over the best model by \citet{whatwhereae}, 
setting the SC as best model at $81.3\%$ test classification accuracy. We also compare with the same
network, but without SC initialization, which achieves a lower classification of $72.6\%$. 
This is an indication that indeed SC managed to leverage unlabeled examples to provide a better
initialization point for the supervised model.

\begin{table}[t]
  \caption{State of the art results on STL-10 dataset}
  \label{stl10_table}
  \centering
  \begin{tabular}{lll}
    \toprule
    \bf Model     & \bf STL-10 test accuracy \\
    \midrule
 Zero-bias Convnets - \citet{paine2014analysis}                        &$70.2\%$\\
 Triplet network - \citet{hoffer2015deep} 	                       &$70.7 \%$ \\
 Exemplar Convnets - \citet{dosovitskiy2014discriminative}             &$72.8\%$ \\
 Target Coding - \citet{yang2015deep}				       &$73.15\%$ \\
 Stacked what-where AE - \citet{whatwhereae}			       &$74.33\%$ \\
   \midrule
Spatial contrasting initialization (this work)                           	    &\bf $81.34\% \pm 0.1$ \\          
The same model without initialization                        	 		   &\bf $72.6\% \pm 0.1$ \\     
    \bottomrule
  \end{tabular}
\end{table}

\subsection{Results on Cifar10}
For Cifar10, we used a previously used setting  \citet{coates2012learning} \citet{hui2013direct}  \citet{dosovitskiy2014discriminative} to test a model's ability to learn from
unlabeled images. In this setting, only $4,000$ samples from the available $50,000$ are used with their label annotation, but the entire dataset is used for unsupervised learning.
The final test accuracy is measured on the entire $10,000$ test set. \\
In our experiments, we trained our model using SC criterion on the entire dataset, and then used only $400$ labeled samples per class (for a total of $4000$) in a supervised regime over 
the initialized network. The results are compared with previous efforts in table \ref{cifar10_table}. Using the SC criterion allowed an improvement of ~6.8\% over a non-initialized model, 
and achieved a final test accuracy of 79.2\%. This is a competitive result with current state-of-the-art model of \citet{rasmus2015semi}. 

      \begin{table}[t]
  \caption{State of the art results on Cifar10 dataset with only 4000 labeled samples}
  \label{cifar10_table}
  \centering
  \begin{tabular}{lll}
    \toprule
    \bf Model     & \bf Cifar10 (400 per class) test accuracy \\
    \midrule
 Convolutional K-means Network - \citet{coates2012learning}                  &$70.7\%$\\
 View-Invariant K-means - \citet{hui2013direct}                            &$72.6\%$\\
 DCGAN - \citet{radford2015unsupervised} 	                      		 &$73.8 \%$ \\
 Exemplar Convnets - \citet{dosovitskiy2014discriminative}             &$76.6\%$ \\
 Ladder networks - \citet{rasmus2015semi}				       &$79.6\%$ \\

   \midrule
Spatial contrasting initialization (this work)                         &\bf $79.2\% \pm 0.3$ \\          
The same model without initialization                        	       &\bf $72.4\% \pm 0.1$ \\     
    \bottomrule
  \end{tabular}
\end{table}

\subsection{Results on MNIST}
The MNIST dataset is very different in nature from the Cifar10 and STL10, we experimented earlier. The biggest difference, relevant to this work, is that spatial regions sampled from MNIST images usually provide very little, or no information. Because of this fact, SC is much less suited for use with MNIST, and was conjured to have little benefit. We still, however, experimented with initializing a model with SC criterion and continuing with a fully-supervised regime over all labeled examples. We found again that this provided benefit over training the same network without pre-initialization, improving results from $0.63\%$ to $0.34\%$ error on test set. The results, compared with previous attempts are included in  \ref{mnist_table}.
\begin{table}[t]
  \caption{results on MNIST dataset}
  \label{mnist_table}
  \centering
  \begin{tabular}{lll}
    \toprule
    \bf Model     & \bf MNIST test error \\
    \midrule
 Stacked what-where AE - \citet{whatwhereae}			       &$0.71\%$ \\
  Triplet network - \citet{hoffer2015deep} 	  &$0.56\% $ \\
  \citet{jarrett2009best} &$0.53\%$\\
  Ladder networks - \citet{rasmus2015semi}				       &$0.36\%$ \\
 DropConnect - \citet{wan2013regularization}						&$0.21\%$\\
   \midrule
Spatial contrasting initialization (this work)                           	    &\bf $0.34\% \pm 0.02$ \\          
The same model without initialization                        	 		   &\bf $0.63\% \pm 0.02$ \\     
    \bottomrule
  \end{tabular}
\end{table}

\section{Conclusions and future work}
In this work we presented spatial contrasting - a novel unsupervised criterion for training convolutional networks on unlabeled data. Its is based on comparison between spatial features
sampled from a number of images. We've shown empirically that using spatial contrasting as a pretraining technique to initialize a ConvNet, 
can improve its performance on a subsequent supervised training. 
In cases where a lot of unlabeled data is available, such as the STL10 dataset, this translates to state-of-the-art classification accuracy in the final model.

Since the spatial contrasting loss is a differentiable estimation that can be computed within a network in parallel to supervised losses,
future work will attempt to embed it as a semi-supervised model. This usage will allow to create models that can leverage
both labeled an unlabeled data, and can be compared to similar semi-supervised models such as the ladder network \citet{rasmus2015semi}.
It is is also apparent that contrasting can occur in dimensions other than the spatial, the most straightforward is the temporal one. This suggests that similar
training procedure can be applied on segments of sequences to learn useful representation without explicit supervision.

\bibliography{spatialcontrasting}

\begin{thebibliography}{36}
\providecommand{\natexlab}[1]{#1}
\providecommand{\url}[1]{\texttt{#1}}
\expandafter\ifx\csname urlstyle\endcsname\relax
  \providecommand{\doi}[1]{doi: #1}\else
  \providecommand{\doi}{doi: \begingroup \urlstyle{rm}\Url}\fi

\bibitem[Balntas et~al.(2016)Balntas, Johns, Tang, and
  Mikolajczyk]{balntas2016pn}
Vassileios Balntas, Edward Johns, Lilian Tang, and Krystian Mikolajczyk.
\newblock Pn-net: Conjoined triple deep network for learning local image
  descriptors.
\newblock \emph{arXiv preprint arXiv:1601.05030}, 2016.

\bibitem[Bromley et~al.(1993)Bromley, Bentz, Bottou, Guyon, LeCun, Moore,
  S{\"a}ckinger, and Shah]{bromley1993signature}
Jane Bromley, James~W Bentz, L{\'e}on Bottou, Isabelle Guyon, Yann LeCun, Cliff
  Moore, Eduard S{\"a}ckinger, and Roopak Shah.
\newblock Signature verification using a “siamese” time delay neural
  network.
\newblock \emph{International Journal of Pattern Recognition and Artificial
  Intelligence}, 7\penalty0 (04):\penalty0 669--688, 1993.

\bibitem[Chopra et~al.(2005)Chopra, Hadsell, and LeCun]{chopra2005learning}
Sumit Chopra, Raia Hadsell, and Yann LeCun.
\newblock Learning a similarity metric discriminatively, with application to
  face verification.
\newblock In \emph{Computer Vision and Pattern Recognition, 2005. CVPR 2005.
  IEEE Computer Society Conference on}, volume~1, pp.\  539--546. IEEE, 2005.

\bibitem[Coates \& Ng(2012)Coates and Ng]{coates2012learning}
Adam Coates and Andrew~Y Ng.
\newblock Learning feature representations with k-means.
\newblock In \emph{Neural Networks: Tricks of the Trade}, pp.\  561--580.
  Springer, 2012.

\bibitem[Coates et~al.(2011)Coates, Ng, and Lee]{stl10}
Adam Coates, Andrew~Y Ng, and Honglak Lee.
\newblock An analysis of single-layer networks in unsupervised feature
  learning.
\newblock In \emph{International Conference on Artificial Intelligence and
  Statistics}, pp.\  215--223, 2011.

\bibitem[Doersch et~al.(2015)Doersch, Gupta, and
  Efros]{doersch2015unsupervised}
Carl Doersch, Abhinav Gupta, and Alexei~A Efros.
\newblock Unsupervised visual representation learning by context prediction.
\newblock In \emph{Proceedings of the IEEE International Conference on Computer
  Vision}, pp.\  1422--1430, 2015.

\bibitem[Dosovitskiy et~al.(2014)Dosovitskiy, Springenberg, Riedmiller, and
  Brox]{dosovitskiy2014discriminative}
Alexey Dosovitskiy, Jost~Tobias Springenberg, Martin Riedmiller, and Thomas
  Brox.
\newblock Discriminative unsupervised feature learning with convolutional
  neural networks.
\newblock In \emph{Advances in Neural Information Processing Systems}, pp.\
  766--774, 2014.

\bibitem[Goodfellow et~al.(2014)Goodfellow, Pouget-Abadie, Mirza, Xu,
  Warde-Farley, Ozair, Courville, and Bengio]{goodfellow2014generative}
Ian Goodfellow, Jean Pouget-Abadie, Mehdi Mirza, Bing Xu, David Warde-Farley,
  Sherjil Ozair, Aaron Courville, and Yoshua Bengio.
\newblock Generative adversarial nets.
\newblock In \emph{Advances in Neural Information Processing Systems}, pp.\
  2672--2680, 2014.

\bibitem[Gutmann \& Hyv{\"a}rinen(2010)Gutmann and Hyv{\"a}rinen]{nce1}
Michael Gutmann and Aapo Hyv{\"a}rinen.
\newblock Noise-contrastive estimation: A new estimation principle for
  unnormalized statistical models.
\newblock In \emph{International Conference on Artificial Intelligence and
  Statistics}, pp.\  297--304, 2010.

\bibitem[He et~al.(2015)He, Zhang, Ren, and Sun]{res}
Kaiming He, Xiangyu Zhang, Shaoqing Ren, and Jian Sun.
\newblock Deep residual learning for image recognition.
\newblock \emph{arXiv preprint arXiv:1512.03385}, 2015.

\bibitem[Hinton(2007)]{hinton2007recognize}
Geoffrey~E Hinton.
\newblock To recognize shapes, first learn to generate images.
\newblock \emph{Progress in brain research}, 165:\penalty0 535--547, 2007.

\bibitem[Hoffer \& Ailon(2015)Hoffer and Ailon]{hoffer2015deep}
Elad Hoffer and Nir Ailon.
\newblock Deep metric learning using triplet network.
\newblock In \emph{Similarity-Based Pattern Recognition}, pp.\  84--92.
  Springer, 2015.

\bibitem[Hui(2013)]{hui2013direct}
Ka~Y Hui.
\newblock Direct modeling of complex invariances for visual object features.
\newblock In \emph{Proceedings of the 30th International Conference on Machine
  Learning (ICML-13)}, pp.\  352--360, 2013.

\bibitem[Ioffe \& Szegedy(2015)Ioffe and Szegedy]{ioffe2015batch}
Sergey Ioffe and Christian Szegedy.
\newblock Batch normalization: Accelerating deep network training by reducing
  internal covariate shift.
\newblock In \emph{Proceedings of The 32nd International Conference on Machine
  Learning}, pp.\  448--456, 2015.

\bibitem[Jarrett et~al.(2009)Jarrett, Kavukcuoglu, Ranzato, and
  LeCun]{jarrett2009best}
Kevin Jarrett, Koray Kavukcuoglu, Marc'Aurelio Ranzato, and Yann LeCun.
\newblock What is the best multi-stage architecture for object recognition?
\newblock In \emph{Computer Vision, 2009 IEEE 12th International Conference
  on}, pp.\  2146--2153. IEEE, 2009.

\bibitem[Krizhevsky \& Hinton(2009)Krizhevsky and
  Hinton]{krizhevsky2009learning}
Alex Krizhevsky and Geoffrey Hinton.
\newblock Learning multiple layers of features from tiny images.
\newblock \emph{Computer Science Department, University of Toronto, Tech. Rep},
  2009.

\bibitem[Krizhevsky et~al.(2012)Krizhevsky, Sutskever, and
  Hinton]{Krizhevsky2012}
Alex Krizhevsky, Ilya Sutskever, and Geoffrey~E Hinton.
\newblock {ImageNet Classification with Deep Convolutional Neural Networks}.
\newblock \emph{Advances In Neural Information Processing Systems}, pp.\  1--9,
  2012.

\bibitem[LeCun et~al.(1998)LeCun, Bottou, Bengio, and
  Haffner]{lecun1998gradient}
Yann LeCun, L{\'e}on Bottou, Yoshua Bengio, and Patrick Haffner.
\newblock Gradient-based learning applied to document recognition.
\newblock \emph{Proceedings of the IEEE}, 86\penalty0 (11):\penalty0
  2278--2324, 1998.

\bibitem[Lin et~al.(2013)Lin, Chen, and Yan]{nin}
Min Lin, Qiang Chen, and Shuicheng Yan.
\newblock Network in network.
\newblock \emph{arXiv preprint arXiv:1312.4400}, 2013.

\bibitem[Mikolov et~al.(2013)Mikolov, Sutskever, Chen, Corrado, and
  Dean]{word2vec}
Tomas Mikolov, Ilya Sutskever, Kai Chen, Greg~S Corrado, and Jeff Dean.
\newblock Distributed representations of words and phrases and their
  compositionality.
\newblock In \emph{Advances in neural information processing systems}, pp.\
  3111--3119, 2013.

\bibitem[Mnih \& Kavukcuoglu(2013)Mnih and Kavukcuoglu]{nce2}
Andriy Mnih and Koray Kavukcuoglu.
\newblock Learning word embeddings efficiently with noise-contrastive
  estimation.
\newblock In \emph{Advances in Neural Information Processing Systems}, pp.\
  2265--2273, 2013.

\bibitem[Mnih et~al.(2015)Mnih, Kavukcuoglu, Silver, Rusu, Veness, Bellemare,
  Graves, Riedmiller, Fidjeland, Ostrovski, et~al.]{mnih2015human}
Volodymyr Mnih, Koray Kavukcuoglu, David Silver, Andrei~A Rusu, Joel Veness,
  Marc~G Bellemare, Alex Graves, Martin Riedmiller, Andreas~K Fidjeland, Georg
  Ostrovski, et~al.
\newblock Human-level control through deep reinforcement learning.
\newblock \emph{Nature}, 518\penalty0 (7540):\penalty0 529--533, 2015.

\bibitem[Ng(2011)]{ng2011sparse}
Andrew Ng.
\newblock Sparse autoencoder.
\newblock 2011.

\bibitem[Paine et~al.(2014)Paine, Khorrami, Han, and Huang]{paine2014analysis}
Tom~Le Paine, Pooya Khorrami, Wei Han, and Thomas~S Huang.
\newblock An analysis of unsupervised pre-training in light of recent advances.
\newblock \emph{arXiv preprint arXiv:1412.6597}, 2014.

\bibitem[Pinheiro et~al.(2015)Pinheiro, Collobert, and
  Dollar]{pinheiro2015learning}
Pedro~O Pinheiro, Ronan Collobert, and Piotr Dollar.
\newblock Learning to segment object candidates.
\newblock In \emph{Advances in Neural Information Processing Systems}, pp.\
  1981--1989, 2015.

\bibitem[Radford et~al.(2015)Radford, Metz, and
  Chintala]{radford2015unsupervised}
Alec Radford, Luke Metz, and Soumith Chintala.
\newblock Unsupervised representation learning with deep convolutional
  generative adversarial networks.
\newblock \emph{arXiv preprint arXiv:1511.06434}, 2015.

\bibitem[Rasmus et~al.(2015)Rasmus, Berglund, Honkala, Valpola, and
  Raiko]{rasmus2015semi}
Antti Rasmus, Mathias Berglund, Mikko Honkala, Harri Valpola, and Tapani Raiko.
\newblock Semi-supervised learning with ladder networks.
\newblock In \emph{Advances in Neural Information Processing Systems}, pp.\
  3532--3540, 2015.

\bibitem[Razavian et~al.(2014)Razavian, Azizpour, Sullivan, and
  Carlsson]{razavian2014cnn}
Ali Razavian, Hossein Azizpour, Josephine Sullivan, and Stefan Carlsson.
\newblock Cnn features off-the-shelf: an astounding baseline for recognition.
\newblock In \emph{Proceedings of the IEEE Conference on Computer Vision and
  Pattern Recognition Workshops}, pp.\  806--813, 2014.

\bibitem[Rumelhart et~al.()Rumelhart, Hinton, and
  Williams]{rumelhart1988learning}
David~E Rumelhart, Geoffrey~E Hinton, and Ronald~J Williams.
\newblock Learning representations by back-propagating errors.
\newblock \emph{Cognitive modeling}, 5\penalty0 (3):\penalty0 1.

\bibitem[Schroff et~al.(2015)Schroff, Kalenichenko, and
  Philbin]{schroff2015facenet}
Florian Schroff, Dmitry Kalenichenko, and James Philbin.
\newblock Facenet: A unified embedding for face recognition and clustering.
\newblock In \emph{Proceedings of the IEEE Conference on Computer Vision and
  Pattern Recognition}, pp.\  815--823, 2015.

\bibitem[Szegedy et~al.(2015)Szegedy, Liu, Jia, Sermanet, Reed, Anguelov,
  Erhan, Vanhoucke, and Rabinovich]{inception}
Christian Szegedy, Wei Liu, Yangqing Jia, Pierre Sermanet, Scott Reed, Dragomir
  Anguelov, Dumitru Erhan, Vincent Vanhoucke, and Andrew Rabinovich.
\newblock Going deeper with convolutions.
\newblock In \emph{Proceedings of the IEEE Conference on Computer Vision and
  Pattern Recognition}, pp.\  1--9, 2015.

\bibitem[Vincent et~al.(2008)Vincent, Larochelle, Bengio, and
  Manzagol]{vincent2008extracting}
Pascal Vincent, Hugo Larochelle, Yoshua Bengio, and Pierre-Antoine Manzagol.
\newblock Extracting and composing robust features with denoising autoencoders.
\newblock In \emph{Proceedings of the 25th international conference on Machine
  learning}, pp.\  1096--1103. ACM, 2008.

\bibitem[Wan et~al.(2013)Wan, Zeiler, Zhang, Cun, and
  Fergus]{wan2013regularization}
Li~Wan, Matthew Zeiler, Sixin Zhang, Yann~L Cun, and Rob Fergus.
\newblock Regularization of neural networks using dropconnect.
\newblock In \emph{Proceedings of the 30th International Conference on Machine
  Learning (ICML-13)}, pp.\  1058--1066, 2013.

\bibitem[Yang et~al.(2015)Yang, Luo, Loy, Shum, and Tang]{yang2015deep}
Shuo Yang, Ping Luo, Chen~Change Loy, Kenneth~W Shum, and Xiaoou Tang.
\newblock Deep representation learning with target coding.
\newblock 2015.

\bibitem[Zeiler et~al.(2010)Zeiler, Krishnan, Taylor, and
  Fergus]{zeiler2010deconvolutional}
Matthew~D Zeiler, Dilip Krishnan, Graham~W Taylor, and Rob Fergus.
\newblock Deconvolutional networks.
\newblock In \emph{Computer Vision and Pattern Recognition (CVPR), 2010 IEEE
  Conference on}, pp.\  2528--2535. IEEE, 2010.

\bibitem[Zhao et~al.(2015)Zhao, Mathieu, Goroshin, and Lecun]{whatwhereae}
Junbo Zhao, Michael Mathieu, Ross Goroshin, and Yann Lecun.
\newblock Stacked what-where auto-encoders.
\newblock \emph{arXiv preprint arXiv:1506.02351}, 2015.

\end{thebibliography}
\bibliographystyle{iclr2017_conference}

\section{Appendix}
\begin{table}[h]
\caption{Convolutional models used, based on \citet{nin},  \citet{rasmus2015semi}}
\label{table:conv-models}
\begin{center}
\begin{small}
\begin{tabular}{l|l|l}
\multicolumn{3}{c}{\bf Model} \\
\hline
STL10         &   CIFAR-10 & MNIST \\
\hline
Input: $96\times 96$ RGB &Input:  $32 \times 32$ RGB &Input:  $28 \times 28$ monochrome\\
\hline
$5 \times 5$ conv. $64$ BN ReLU  & $3 \times 3$ conv. $96$ BN LeakyReLU  & $5 \times 5$ conv. $32$ ReLU  \\
$1 \times 1$ conv. $160$ BN ReLU  & $3 \times 3$ conv. $96$ BN LeakyReLU  &   \\
$1 \times 1$ conv. $96$ BN ReLU  & $3 \times 3$ conv. $96$ BN LeakyReLU  &   \\
$3 \times 3$ max-pooling, stride $2$  & $2 \times 2$ max-pooling, stride $2$ BN & $2 \times 2$ max-pooling, stride $2$ BN  \\
$5 \times 5$ conv. $192$ BN ReLU & $3 \times 3$ conv. $192$ BN LeakyReLU & $3 \times 3$ conv. $64$ BN ReLU \\
$1 \times 1$ conv. $192$ BN ReLU & $3 \times 3$ conv. $192$ BN LeakyReLU & $3 \times 3$ conv. $64$ BN ReLU \\
$1 \times 1$ conv. $192$ BN ReLU & $3 \times 3$ conv. $192$ BN LeakyReLU &  \\
$3 \times 3$ max-pooling, stride $2$  & $2 \times 2$ max-pooling, stride $2$ BN & $2 \times 2$ max-pooling, stride $2$ BN  \\
$3 \times 3$ conv. $192$ BN ReLU & \\
$1 \times 1$ conv. $192$ BN ReLU & \\
$1 \times 1$ conv. $192$ BN ReLU & \\
\hline
\multicolumn{3}{c}{Spatial contrasting criterion} \\
\hline

$3 \times 3$ conv. $256$ ReLU & $3 \times 3$ conv. $192$ BN LeakyReLU &  $3 \times 3$ conv. $128$ BN ReLU \\
$3 \times 3$ max-pooling, stride $2$  & $1 \times 1$ conv. $192$ BN LeakyReLU &  $1 \times 1$ conv. $10$ BN ReLU \\
dropout, $p=0.5$  &  $1 \times 1$ conv. $10$  BN LeakyReLU &    global average pooling \\
$3 \times 3$ conv. $128$ ReLU &   global average pooling&  \\
dropout, $p=0.5$  &  &  \\
fully-connected $10$ & & \\
\hline
\multicolumn{3}{c}{10-way softmax} \\
\end{tabular}
\end{small}
\end{center}
\end{table}

\begin{figure}[h]
  \centering
  \label{stl10_filters}
\includegraphics[width=0.4\textwidth]{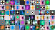}
  \caption{First layer convolutional filters after spatial-contrasting training}
\end{figure}
\end{document}